\documentclass{llncs}
\usepackage[T1]{fontenc}
\usepackage{graphicx}
\usepackage[colorlinks=true, urlcolor=blue, citecolor=black, linkcolor=black]{hyperref}

\usepackage{tikz}
\usetikzlibrary{decorations.pathreplacing}
\usepackage{amsmath}
\usepackage{amssymb}
\usepackage{booktabs}
\usepackage{caption}
\usepackage{subcaption}
\usepackage{adjustbox}

\begin{document}
\title{Car Drag Coefficient Prediction from 3D Point
Clouds Using a Slice-Based Surrogate Model}

\author{Utkarsh Singh\inst{1}\orcidID{0009-0003-6327-8392} \and
Absaar Ali\inst{1}\orcidID{0009-0004-1408-6690} \and
Adarsh Roy\inst{2}\orcidID{0009-0009-4966-0660}}

\institute{Delhi Technological University, Shahbad Daulatpur, Delhi, 110042,
India\\
\email{\{utkarshsingh\_me21b16\_52,absaarali\_co20b2\_24\}@dtu.ac.in}
\and
Indian Institute of Technology, Hauz
Khas, Delhi, 110016, India\\
\email{adarsh.roy@iitdalumni.com}
}

\maketitle

\begin{abstract}
The automotive industry's pursuit of enhanced fuel economy and performance necessitates efficient aerodynamic design. However, traditional evaluation methods such as computational fluid dynamics (CFD) and wind tunnel testing are resource intensive, hindering rapid iteration in the early design stages. Machine learning-based surrogate models offer a promising alternative, yet many existing approaches suffer from high computational complexity, limited interpretability, or insufficient accuracy for detailed geometric inputs. This paper introduces a novel lightweight surrogate model for the prediction of the aerodynamic drag coefficient ($C_d$) based on a sequential slice-wise processing of the geometry of the 3D vehicle. Inspired by medical imaging, 3D point clouds of vehicles are decomposed into an ordered sequence of 2D cross-sectional slices along the stream-wise axis. Each slice is encoded by a lightweight PointNet2D module, and the sequence of slice embeddings is processed by a bidirectional LSTM to capture longitudinal geometric evolution. The model, trained and evaluated on the DrivAerNet++ dataset, achieves a high coefficient of determination ($R^2 > 0.9528$) and a low mean absolute error ($MAE \approx 6.046 \times 10^{-3}$) in $C_d$ prediction. With an inference time of approximately 0.025 seconds per sample on a consumer-grade GPU, our approach provides fast, accurate, and interpretable aerodynamic feedback, facilitating more agile and informed automotive design exploration.

\keywords{Drag Coefficient Prediction \and Automotive Aerodynamics \and Surrogate Modeling \and Slice-Based Model \and Point Clouds \and Deep Learning}
\end{abstract}

\section{Introduction}\label{sec1}
Aerodynamic efficiency is paramount in the automotive industry, directly impacting fuel economy, emissions, vehicle stability, and the range of electric vehicles. Reducing aerodynamic drag, quantified by the drag coefficient ($C_d$), is a primary design goal. However, conventional evaluation methods, namely Computational Fluid Dynamics (CFD) and wind tunnel testing, present significant bottlenecks. CFD simulations, while detailed, are computationally expensive and time consuming, with typical runs taking hours to days and requiring substantial high-performance computing (HPC) resources \cite{ferrari2023automated,hpc_costs2025}. Wind tunnel tests, though crucial for validation, involve costly facility operation and lengthy model fabrication times \cite{ford_wind_tunnel_2017}, limiting their use in the early iterative design phases. These constraints are a hindrance to rapid exploration of the design space.

To overcome these limitations, few machine learning surrogate models have emerged as a promising alternative for rapid aerodynamic prediction \cite{akasaka2022surrogate}. These models learn a complex mapping from vehicle geometry to aerodynamic properties from data generated by high-fidelity simulations. Once trained, they can predict $C_d$ in seconds or milliseconds.
However, existing ML surrogates face several challenges. Voxel-based methods, using 3D Convolutional Neural Networks (CNNs), suffer from resolution bottlenecks and can lose fine geometric details \cite{liu2019pointvoxel}. Projection-based methods, which convert 3D shapes into 2D images for 2D CNNs \cite{song2023data}, can suffer from information loss due to occlusions or choice of viewpoints and may lack physical interpretability. Point cloud-based methods, such as PointNet \cite{qi2017pointnet} and its variants \cite{chen2025tripnet,qi2017pointnetplusplus}, operate directly on 3D surface points but often treat points permutation-invariantly, potentially missing crucial directional cues inherent in aerodynamic flow. More recent graph neural networks \cite{elrefaie2024drivaernet} and transformer-based architectures \cite{he2025drivaer,jiang2023transcfd,xiang2024aerodit}, while powerful, can be computationally intensive and complex. For instance, TripNet \cite{chen2025tripnet} uses triplane representations and achieves high accuracy, but still involves sophisticated geometric processing. Many of these models, particularly complex deep learning architectures, act as "black boxes," offering limited insight into how specific geometric features influence aerodynamic performance.

This paper proposes a novel lightweight, sequential and interpretable approach for $C_d$ prediction. Our core idea is to represent the 3D vehicle geometry as an \textbf{ordered} sequence of 2D cross-sectional slices along the primary (streamwise) direction of airflow, analogous to how MRI or CT scans represent 3D anatomical structures. This structured representation explicitly captures the front-to-rear evolution of the vehicle's shape, which is fundamental to its aerodynamic behavior.
Each 2D slice, represented as a set of points, is processed by our lightweight \textbf{PointNet2D} module (our 2D adaptation of PointNet \cite{qi2017pointnet}) to extract local geometric features. The sequence of these per-slice feature embeddings is then fed into a \textbf{Bidirectional Long Short-Term Memory (LSTM)} network, which models the dependencies and progression of shape features along the vehicle's length. Finally, a Multi-Layer Perceptron (MLP) regresses the $C_d$ from the LSTM's aggregated representation.
This approach offers several advantages :
\begin{itemize}
    \item \textbf{Efficiency:} By processing 2D slices, it avoids the high computational cost of full 3D convolutions or global attention mechanisms on large point clouds.
    \item \textbf{Interpretability:} The model directly captures the flow-aware, front-to-rear progression of the vehicle's shape. This intuitive approach provides a basis for attributing drag contributions to specific longitudinal sections of the vehicle.
\end{itemize}
We validate our model using the large-scale DrivAerNet++ dataset \cite{elrefaie2024drivaernetpp}, which provides high-fidelity CFD-computed $C_d$ values for thousands of parametric car models. Our initial results demonstrate competitive accuracy with state-of-the-art methods, but with significantly reduced computational complexity and enhanced interpretability.

The remainder of this paper is organized as follows: Section~\ref{sec2} details the dataset, data pre-processing techniques, and, the proposed model architecture. Section~\ref{sec3} presents the experimental results, including performance comparisons and training dynamics. Section~\ref{sec4} discusses the implications of these results, limitations, and comparisons. Finally, Section~\ref{sec5} concludes the paper and outlines future research directions.

\section{Methods}\label{sec2}
This section details the dataset, pre-processing steps, the architecture of our proposed slice-based sequential model, and the training procedure.

\subsection{Dataset}
We utilize the DrivAerNet++ dataset \cite{elrefaie2024drivaernetpp}, a large-scale multi-modal car aerodynamics dataset. It contains over 8,000 parametric car models, spanning various body styles (fastback, notchback, estateback, SUV) and configurations (e.g., with/without detailed underbodies, rotating wheels). For each model, the dataset provides a 3D mesh, a point cloud representation, and the ground truth drag coefficient ($C_d$) computed using Reynolds-Averaged Navier-Stokes (RANS) $k-\omega$ SST CFD simulations.
We use the point cloud representation provided via the PaddleScience platform~\cite{paddlescience_docs2025}. After filtering for complete data, our working dataset comprises 7,713 unique car geometries. We adopt the following split: 5398 samples for training, 1115 for validation, and 1200 for testing. This choice ensures a fair comparison with prior studies, which also evaluated on a 1200-sample test set. The point clouds are consistently oriented with the X-axis along the streamwise direction, Y-axis laterally, and Z-axis vertically.

\subsection{Data Pre-processing}
The core of our pre-processing pipeline is the conversion of each 3D car point cloud into an ordered sequence of 2D cross-sectional slices.

\subsubsection{Cross-Sectional Slicing}
\begin{figure}[htbp]
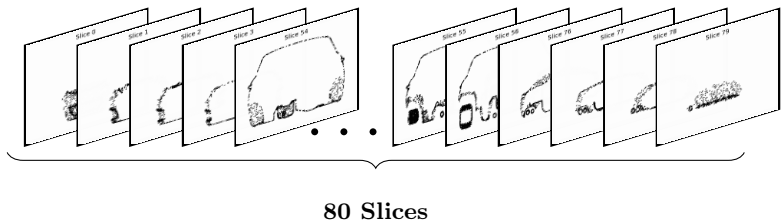

  \centering
  \begin{tikzpicture}[x=1cm,y=1cm]
    \def\step{0.70}
    \def\imgw{1.6cm}
    \foreach \i in {0,...,4}{
       \begin{scope}[cm={1,0.30,0,1,(\i*\step,0)}]
         \node[transform shape,
               draw,black,ultra thick,
               inner sep=0pt]
           {\includegraphics[width=\imgw]{figures/slice_stack/slice_\i.png}};
       \end{scope}
    }
    \begin{scope}[cm={1,0.30,0,1,(5*\step,0)}]
       \node[minimum width=\imgw,minimum height=0pt]{};
       \node at (0,-0.75) {\Huge\bfseries $\cdots$};
    \end{scope}

    \foreach \i in {5,...,10}{
       \pgfmathsetmacro\shift{(\i+2)*\step}
       \begin{scope}[cm={1,0.30,0,1,(\shift,0)}]
         \node[transform shape,
               draw,black,ultra thick,
               inner sep=0pt]
           {\includegraphics[width=\imgw]{figures/slice_stack/slice_\i.png}};
       \end{scope}
    }
    \draw[
      decorate,
      decoration={brace,amplitude=6pt,mirror},
    ] 
      ({-1.5*\step},-1) -- ({(10+2.5)*\step},-1)
      node[midway,yshift=-22pt]{\bfseries 80 Slices};
  \end{tikzpicture}
  \caption{Illustration of 80 streamwise (x-axis) slices extracted from a car’s point cloud. This isometric side view demonstrates how dense slicing captures detailed shape variation from front to back.}
  \label{fig:slice_stack}
\end{figure}

\begin{figure}[htbp]
  \centering
  \includegraphics[width=\linewidth]{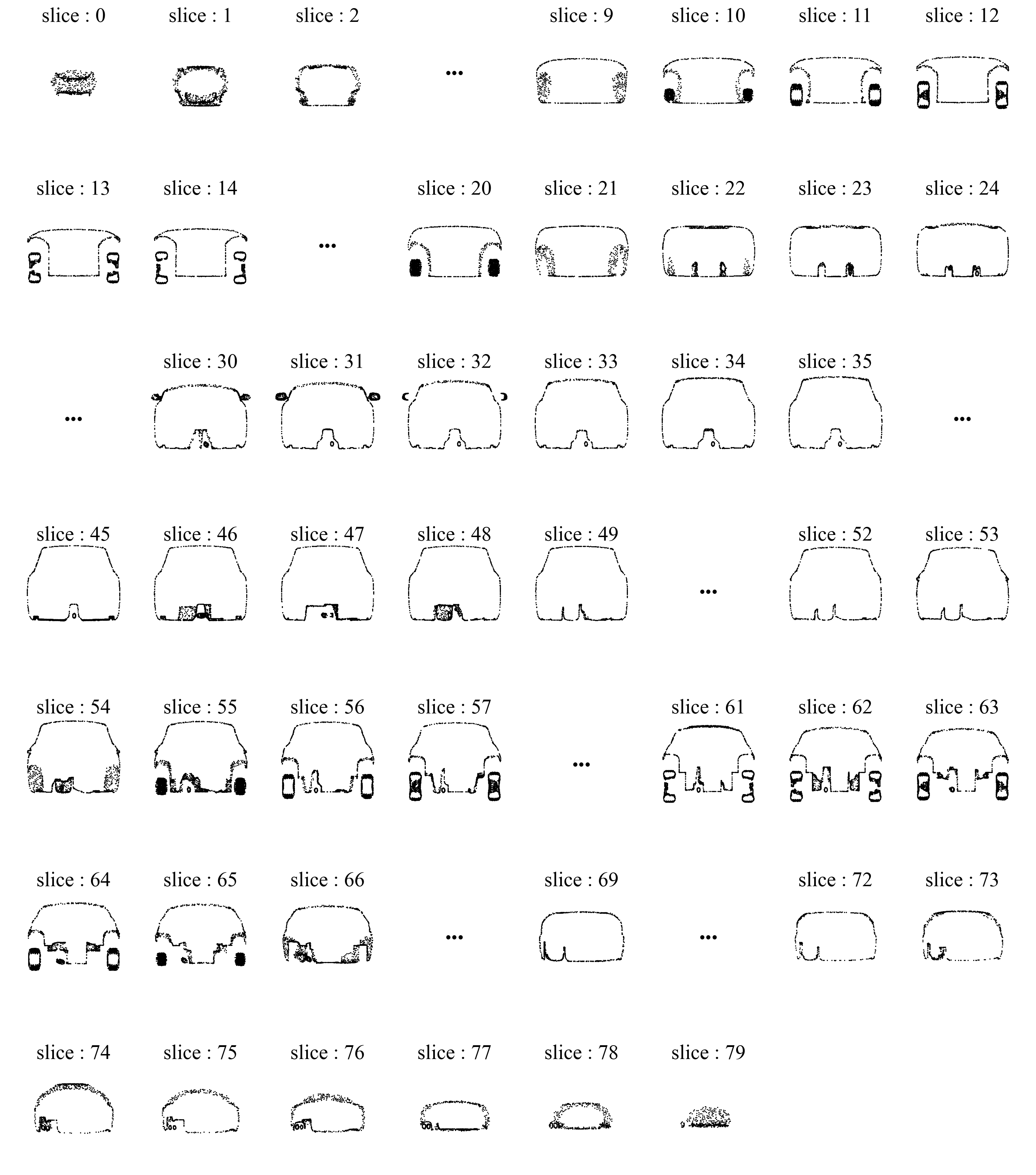}
  \caption{Visualization of 80 streamwise (X-axis) cross-sectional slices extracted from a vehicle's point cloud. This grid displays the individual 2D point sets, sequentially arranged to highlight the progression of the vehicle's contour from front to back, as used for feature extraction.}
  \label{fig:slicing_grid}
\end{figure}

Each 3D point cloud (typically $\sim$100,000 points) is sliced along the primary flow direction ($X$-axis).
\begin{enumerate}
    \item \textbf{Number of Slices ($S$):} We chose $S=80$ slices. This value was empirically found to provide a good balance between capturing sufficient geometric detail along the car's length and maintaining a manageable sequence length for the LSTM. Too few slices would blur important local features, while too many would increase computational cost and redundancy.
    \item \textbf{Binning:} For each car, the range of $x$ coordinates ($x_{min}, x_{max}$) is determined. This range is then divided into $S$ equal bins. The width of each bin $w$ is $(x_{max} - x_{min}) / S$.
    \item \textbf{Projection to $YZ$-Plane:} All points within the $i$-th bin (that is, $x \in [x_{min} + (i-1)w, x_{min} + iw)$) are projected onto the $YZ$-plane by discarding their $x$ coordinate. This results in a set of 2D representing the cross-sectional profile of the car at that longitudinal station.
\end{enumerate}
This process yields a sequence of $S=80$ slices, each represented by a variable number of 2D points $(y,z)$. The overall slicing strategy is conceptually depicted in Figure~\ref{fig:slice_stack}, while Figure~\ref{fig:slicing_grid} visualizes the sequence for a sample vehicle, showcasing the detailed evolution of its cross-sectional shape from front to rear.

\subsubsection{Padding and Masking}
The number of points in each 2D slice varies. To create fixed-size tensors for batch processing, we pad each slice.
\begin{itemize}
    \item We determined the maximum number of points observed in any single slice in the entire dataset, $M_{max}$ (found to be 6,500).
    \item Each 2D slice is zero-padded to have $M_{max}$ points. Thus, each slice becomes a tensor of shape $(M_{max}, 2)$.
    \item A binary mask tensor of shape $(S, M_{max})$ is created to identify real points and ignore padded entries. Although the mask is available, PointNet2D's global max-pooling is generally robust to zero-padded points, provided features are non-negative.
\end{itemize}
The final input representation for each car is a tensor of shape $(S, M_{max}, 2)$, i.e., $(80, 6500, 2)$.

\subsection{Model Architecture}
\begin{figure}[htbp]
    \centering
    \includegraphics[width=\linewidth]{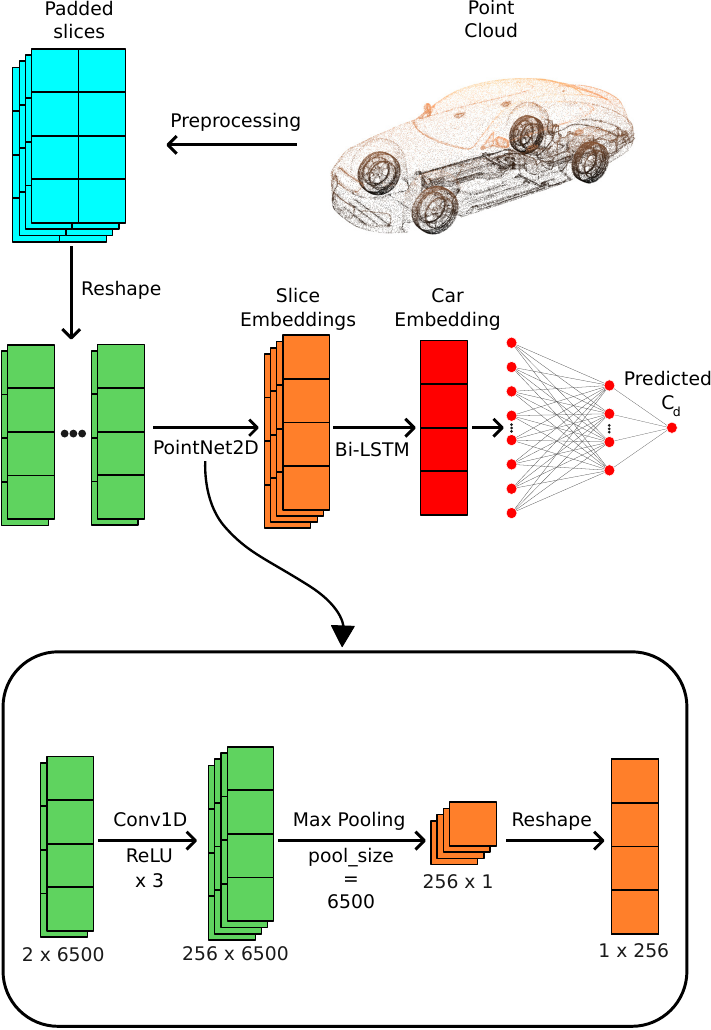}
    \caption{Complete model architecture of the proposed sequential slice-based drag prediction model. 3D point clouds are sliced. Each slice is encoded by PointNet2D. The sequence of embeddings is processed by a Bi-LSTM and an MLP regresses $C_d$.}
    \label{fig:model_architecture_overall}
\end{figure}

Our proposed model consists of three main components: a slice-level feature extractor (PointNet2D), a sequence model (Bi-LSTM), and a regression head (MLP). The complete model architecture is depicted in Figure~\ref{fig:model_architecture_overall}.

\subsubsection{Slice-Level Feature Extraction: PointNet2D}
Each 2D slice (a set of $M_{max}$ points in $\mathbb{R}^2$) is processed independently by our PointNet2D module. This module is a simplified adaptation of the original PointNet \cite{qi2017pointnet} tailored for 2D point sets and designed to be lightweight.
As shown in Figure~\ref{fig:model_architecture_overall}, the PointNet2D module consists of a series of shared 1D convolutional layers (effectively acting as MLPs applied to each point) followed by a global max-pooling operation.
\begin{itemize}
    \item Input: A slice of shape $(M_{max}, 2)$.
    \item Layers: Three 1D convolutional layers with kernel size 1. Channel sizes are $2 \to 32 \to 64 \to d_e=256$. Each convolution is followed by a ReLU activation function.
    \item Max-Pooling: A global max-pooling operation is applied across the $M_{max}$ points dimension to obtain a single feature vector of dimension $d_e=256$ for each slice. This ensures permutation invariance for points within a slice.
\end{itemize}
The output for each car, after this stage, is a sequence of 80 embeddings, resulting in a tensor of shape $(80, 256)$.

\subsubsection{Sequence Modeling: Bi-Directional LSTM}
The sequence of $S=80$ slice embeddings (each of dimension $d_e=256$) is processed by a Bidirectional Long Short-Term Memory (Bi-LSTM) network. The Bi-LSTM captures dependencies and contextual information from both forward (front-to-rear) and backward (rear-to-front) directions of the slice sequence.
\begin{itemize}
    \item Layers: 2 Bi-LSTM layers.
    \item Hidden Dimension: Each LSTM direction has a hidden state dimension of 256.
    \item Output: The final hidden states from the forward and backward passes of the last Bi-LSTM layer are concatenated to form a single feature vector representing the entire vehicle. For a hidden dimension of h=256, this process creates a 512-dimensional car-level embedding.
\end{itemize}

\subsubsection{Regression Head: MLP}
The 512-dimensional car embedding from the Bi-LSTM is fed into a Multi-Layer Perceptron (MLP) to regress the scalar $C_d$ value.
\begin{itemize}
    \item Layers: The MLP consists of three fully connected layers: $512 \to 256 \to 64 \to 1$.
    \item Activations: ReLU activations are used after the first two layers. A dropout layer with a rate of 0.3 is applied after the first ReLU for regularization.
    \item Output: A single scalar value representing the predicted $C_d$.
\end{itemize}
The total number of trainable parameters in the model is approximately 2.79 million.  A summary of the model's component-wise structure, derived directly from its PyTorch implementation, is as follows:
\begin{verbatim}
CdPredictor(
  (pointnet): PointNet2D(
    (slice_encoder): Sequential(
      (0): Conv1d(2, 32, kernel_size=(1,), stride=(1,))
      (1): ReLU(inplace=True)
      (2): Conv1d(32, 64, kernel_size=(1,), stride=(1,))
      (3): ReLU(inplace=True)
      (4): Conv1d(64, 256, kernel_size=(1,), stride=(1,))
      (5): ReLU(inplace=True)
    )
  )
  (sequential_encoder): LSTMSliceEncoder(
    (lstm): LSTM(256, 256, num_layers=2, batch_first=True, 
                 dropout=0.2, bidirectional=True)
  )
  (regressor): CdRegressor(
    (net): Sequential(
      (0): Linear(in_features=512, out_features=256, bias=True)
      (1): ReLU()
      (2): Dropout(p=0.3, inplace=False)
      (3): Linear(in_features=256, out_features=64, bias=True)
      (4): ReLU()
      (5): Linear(in_features=64, out_features=1, bias=True)
    )
  )
)
\end{verbatim}

\subsection{Training Details}
\begin{itemize}
    \item \textbf{Loss Function:} We use the Smooth L1 Loss (Huber Loss), as defined in Equation~\ref{eq:smooth_l1}, which is less sensitive to outliers than Mean Squared Error and provides smooth gradients.
    \begin{equation}
    \label{eq:smooth_l1} 
    \mathcal{L}(y, \hat{y}) =
    \begin{cases}
    0.5 (y - \hat{y})^2, & \text{if } |y - \hat{y}| < \beta \\
    \beta (|y - \hat{y}| - 0.5 \beta), & \text{otherwise}
    \end{cases}
    \end{equation}
    We use $\beta=1.0$.
    \item \textbf{Optimizer:} Adam optimizer with an initial learning rate of $1 \times 10^{-4}$.
    \item \textbf{Batch Size:} A batch size of 4 was used due to GPU memory constraints with the large $M_{max}$.
    \item \textbf{Epochs:} The model was trained for 100 epochs, and the best model was selected based on the highest $R^2$ score on the validation set.
    \item \textbf{Hardware:} A single NVIDIA RTX 4060 Laptop GPU.
\end{itemize}

\section{Results}\label{sec3}
This section presents the performance of our proposed slice-based sequential model. We first define the evaluation metrics, then provide quantitative comparisons with state-of-the-art methods, discuss computational efficiency, and finally analyze training dynamics and error distributions.

\subsection{Evaluation Metrics}
We use standard regression metrics to evaluate model performance:
\begin{itemize}
    \item \textbf{Mean Squared Error (MSE):} $\frac{1}{N}\sum_{i=1}^N(\hat{y}_i - y_i)^2$.\\
    Measures the average squared difference between predicted and true values.
    \item \textbf{Mean Absolute Error (MAE):} $\frac{1}{N}\sum_{i=1}^N|\hat{y}_i - y_i|$.\\
    Measures the average absolute difference, less sensitive to outliers than MSE.
    \item \textbf{Coefficient of Determination ($R^2$):} $1 - \frac{\sum_{i=1}^N(\hat{y}_i - y_i)^2}{\sum_{i=1}^N(y_i - \bar{y})^2}$.\\
    Represents the proportion of variance in the dependent variable that is predictable from the independent variables. An $R^2$ of 1 indicates perfect prediction.
    \item \textbf{Maximum Absolute Error (MaxAE):} $\max_{i}|\hat{y}_i - y_i|$. Indicates the worst-case prediction error.
\end{itemize}
For these metrics, $y_i$ is the true $C_d$, $\hat{y}_i$ is the predicted $C_d$, $\bar{y}$ is the mean of true $C_d$ values, and $N$ is the number of samples.

\subsection{Quantitative Performance}
We compare our model's performance on the DrivAerNet++ test set (1200 samples)
against several published surrogate models. Table~\ref{tab:benchmark_comparison}
summarizes these results. Our PointNet2D + BiLSTM model achieves an $R^2$ of
0.9528 and an MAE of $6.046 \times 10^{-3}$.

\begin{table*}[htbp]
  \small
  \setlength{\tabcolsep}{5pt}
  \centering
  \caption{Quantitative comparison of drag–prediction models on the DrivAerNet++ test set.}
  \label{tab:benchmark_comparison}

  \begin{adjustbox}{width=\textwidth,center}
    \begin{tabular}{@{}lccccc@{}}
      \toprule
      \textbf{Model} & \textbf{Dataset subset} & \textbf{MSE} ($10^{-5}$) &
      \textbf{MAE} ($10^{-3}$) & \textbf{MaxAE} ($10^{-2}$) & $\mathbf{R^{2}}$ \\
      \midrule
      PointNet2D+BiLSTM (Ours) & DrivAerNet++ (1200) &
      \textbf{6.50} & \textbf{6.046} & \textbf{4.50} & 0.9528 \\
      TripNet \cite{chen2025tripnet}                       & DrivAerNet++ (1200) &  9.10 & 7.17 &  7.70 & \textbf{0.957} \\
      RegDGCNN \cite{elrefaie2024drivaernet}\textsuperscript{a}              & DrivAerNet++ (1200) & 14.20 & 9.31 & 12.79 & 0.641 \\
      PointNet \cite{qi2017pointnet}\textsuperscript{a}    & DrivAerNet++ (1200) & 14.90 & 9.60 & 12.45 & 0.643 \\
      \bottomrule
    \end{tabular}
  \end{adjustbox}

  \vspace{2pt}
  \begin{minipage}{\textwidth}
    \footnotesize
    \textsuperscript{a}\,Reported in \cite{elrefaie2024drivaernetpp}. \quad
  \end{minipage}
\end{table*}

Our model demonstrates performance comparable to the state-of-the-art TripNet on DrivAerNet++ in terms of $R^2$, while significantly outperforming earlier methods like RegDGCNN and PointNet. The low MAE indicates that, on average, our predictions are very close to the true $C_d$ values.

\subsection{Computational Efficiency}
Our model is lightweight, with 2.79 million parameters, and achieves the average inference time of 0.025 seconds per sample on an NVIDIA GeForce RTX 4060 Laptop GPU demonstrating the model's suitability for real-time feedback on consumer-grade hardware.

\subsection{Training Dynamics and Error Analysis}
\begin{figure}[htbp]
    \centering
    \begin{subfigure}[b]{0.49\textwidth}
        \includegraphics[width=\textwidth]{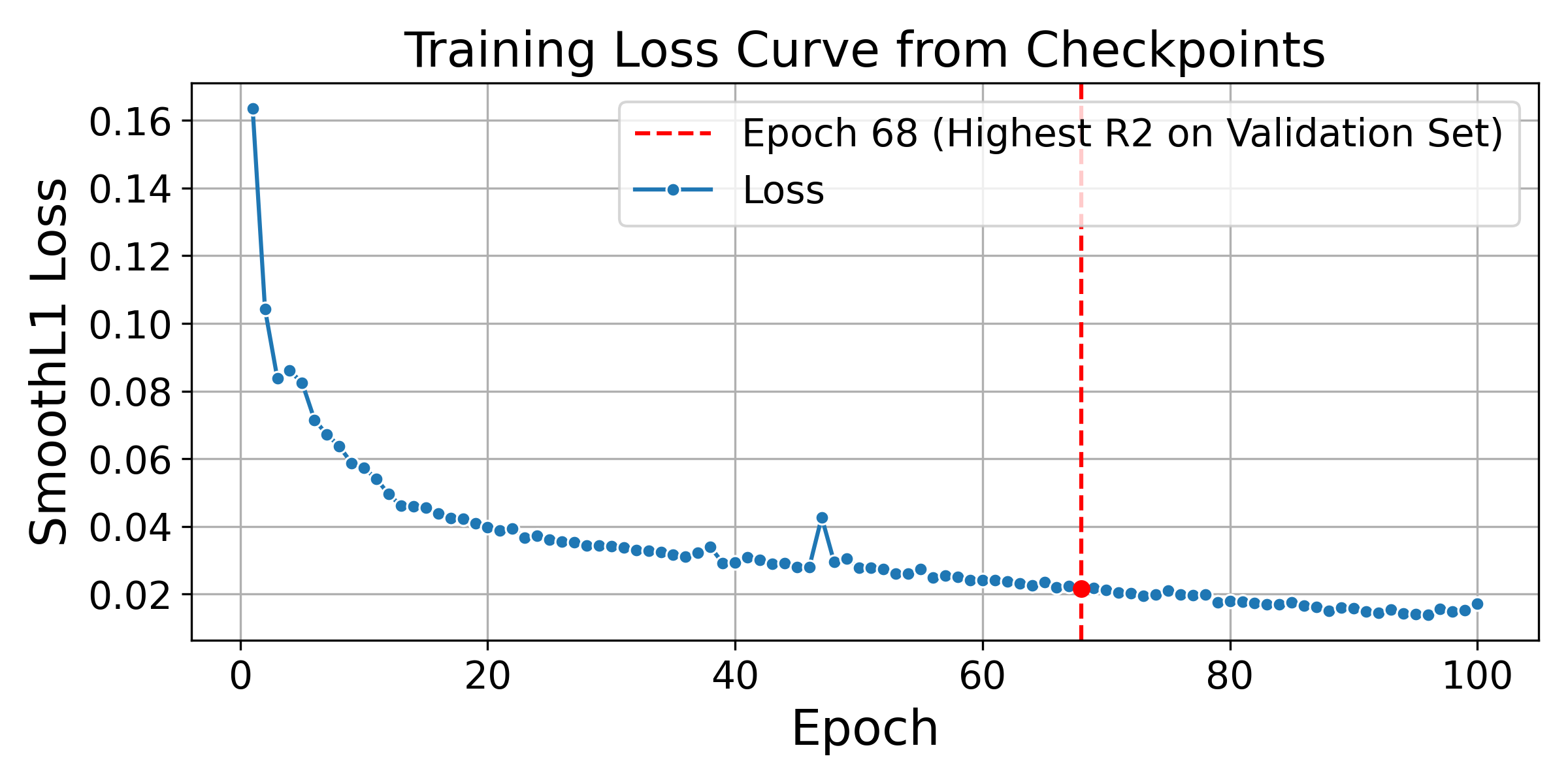}
        \caption{}
        \label{fig:training_loss}
    \end{subfigure}
    \hfill
    \begin{subfigure}[b]{0.49\textwidth}
        \includegraphics[width=\textwidth]{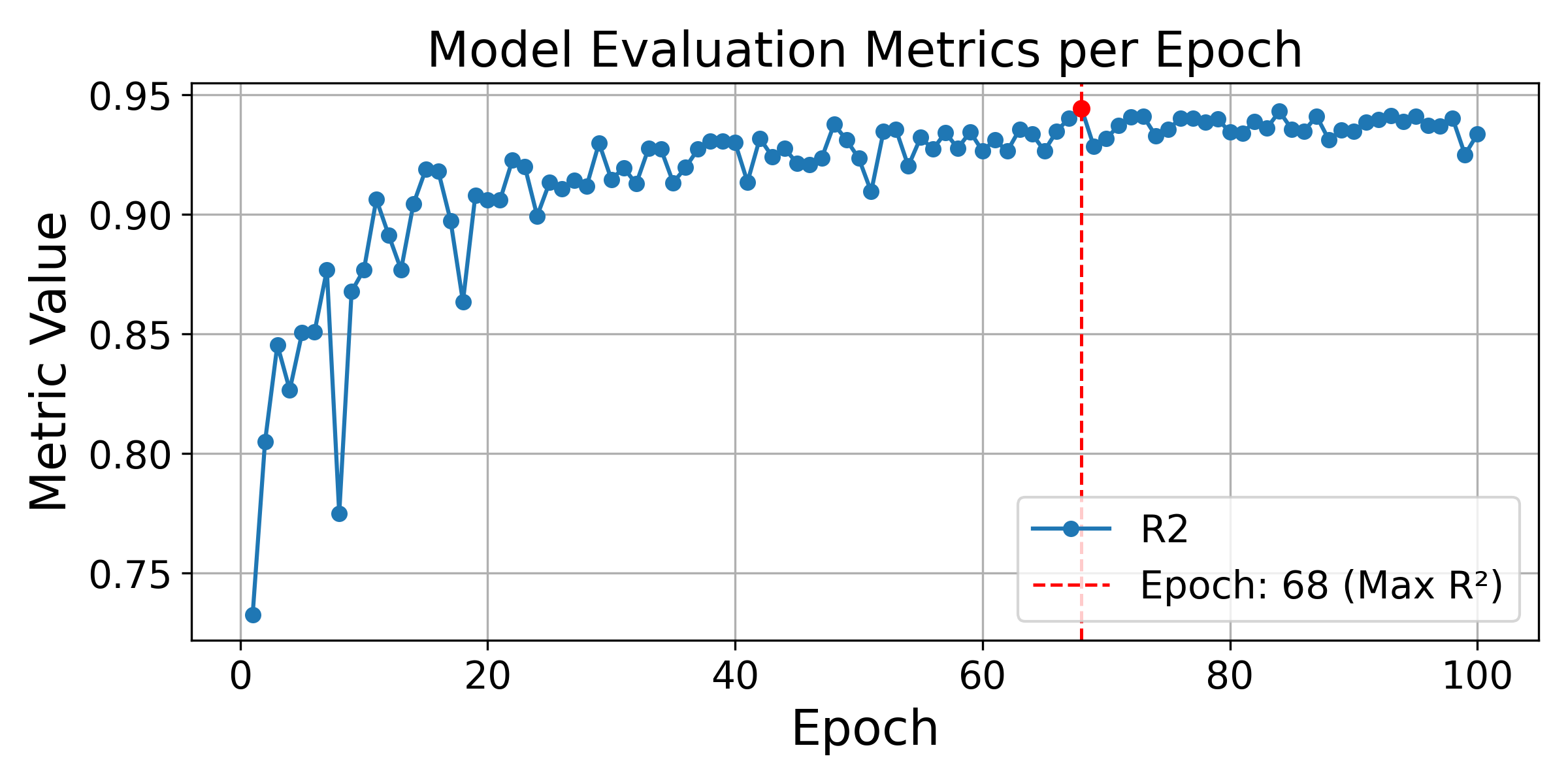}
        \caption{}
        \label{fig:validation_r2}
    \end{subfigure}
    \caption{Training dynamics: (a) Training loss curve. (b) Validation $R^2$ score over epochs. Best validation $R^2$ was achieved at epoch 68.}
\end{figure}

The model was trained for 100 epochs in a single, end-to-end process. Figure~\ref{fig:training_loss} shows the Smooth L1 training loss curve, indicating steady convergence. Figure~\ref{fig:validation_r2} displays the validation $R^2$ score per epoch, with the best performance achieved at epoch 68, which was selected as the final model.

\begin{figure}[htbp]
    \centering
    \begin{subfigure}[b]{0.49\textwidth}
        \includegraphics[width=\textwidth]{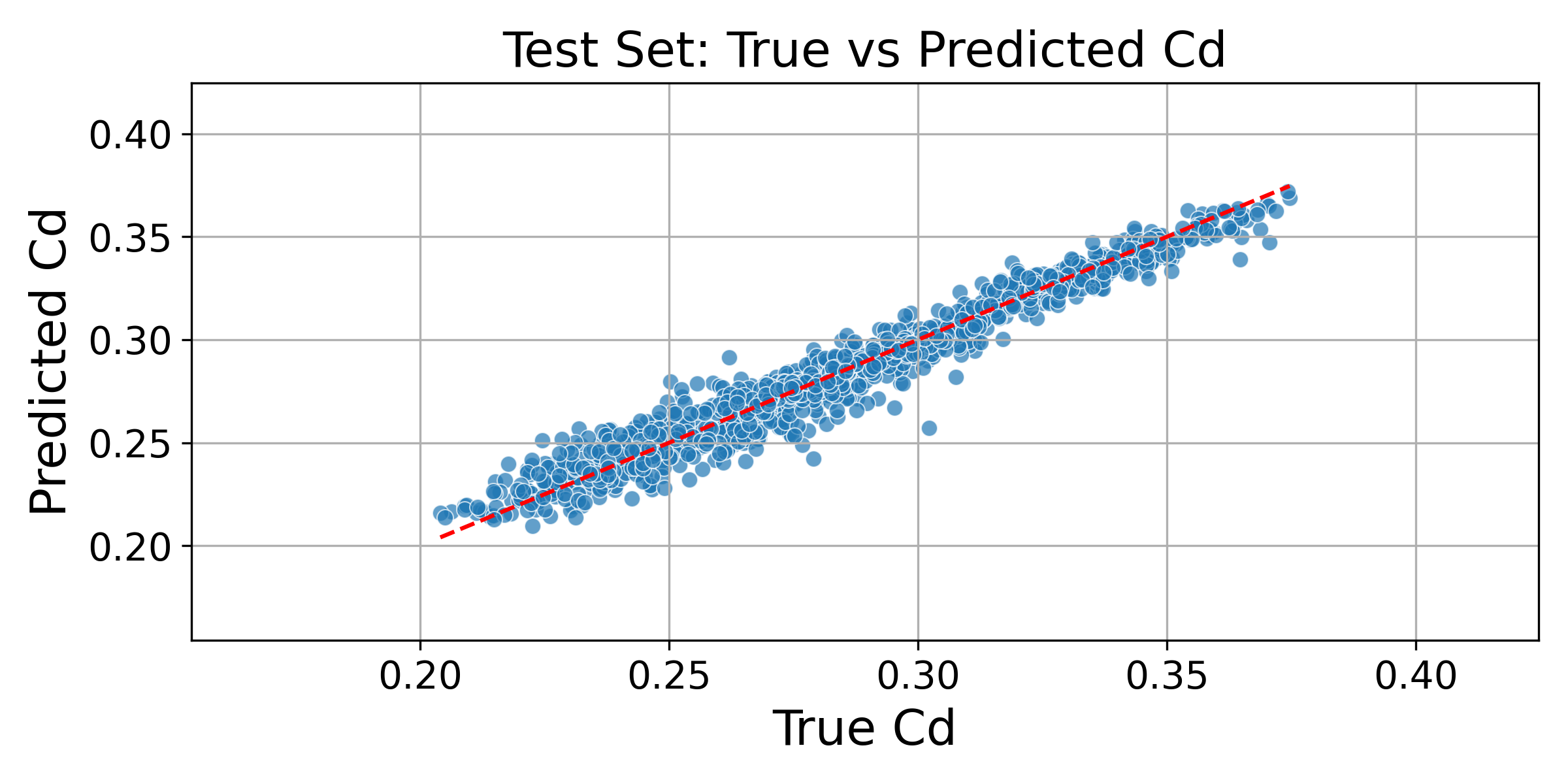}
        \caption{}
        \label{fig:test_scatter}
    \end{subfigure}
    \hfill
    \begin{subfigure}[b]{0.49\textwidth}
        \includegraphics[width=\textwidth]{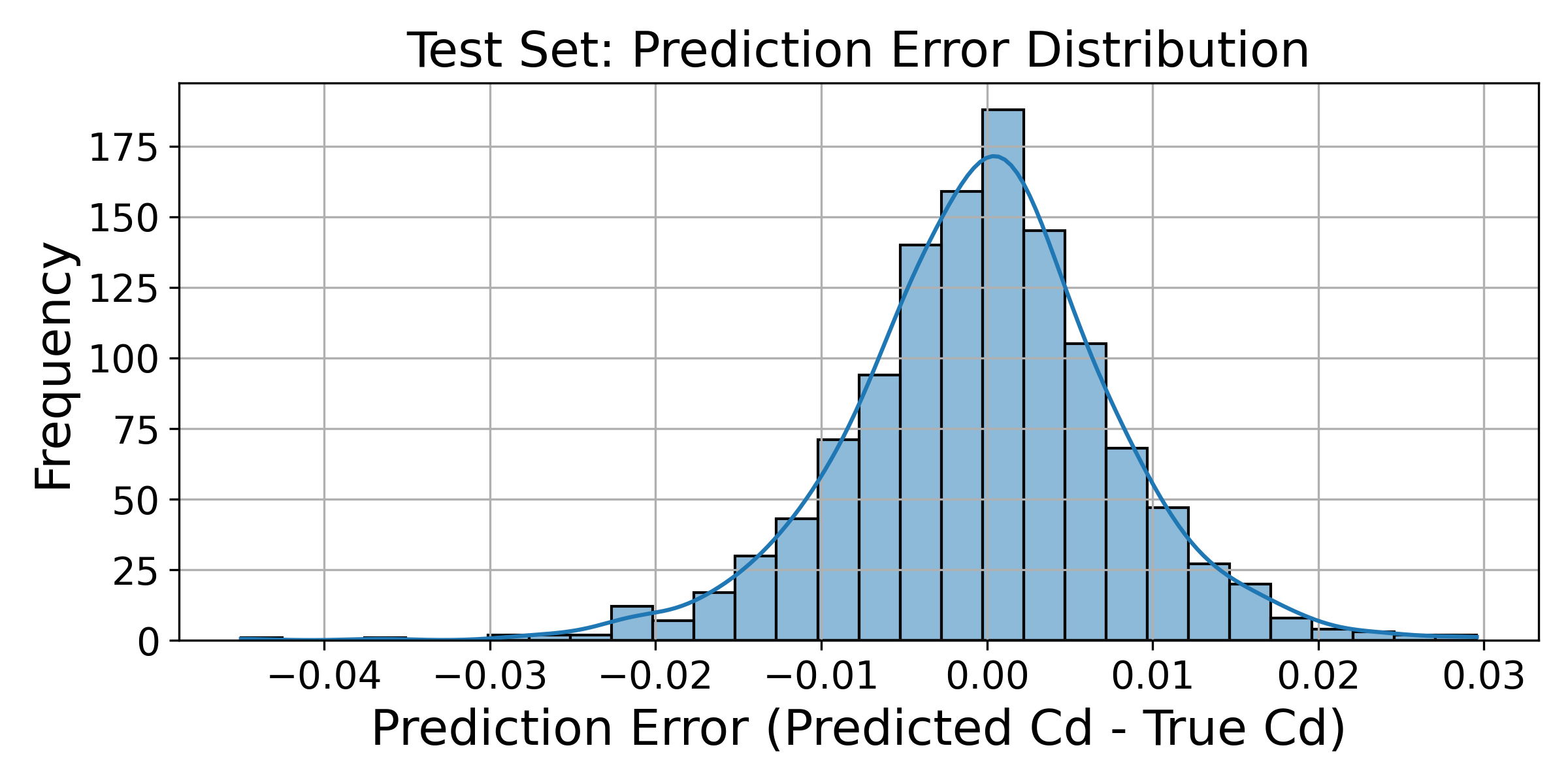}
        \caption{}
        \label{fig:test_error_hist}
    \end{subfigure}
    \caption{Test set performance analysis: (a) Scatter plot showing strong correlation between predicted and true $C_d$ values. (b) Histogram of prediction errors, centered near zero with small spread.}
    \label{fig:test_set_analysis}
\end{figure}

To analyze the prediction quality on the test set, Figure~\ref{fig:test_set_analysis} presents a scatter plot of predicted versus true $C_d$ values and a histogram of prediction errors.

The scatter plot (Figure~\ref{fig:test_scatter}) shows a tight clustering of points around the $y=x$ line, indicating high agreement between the predictions and the ground truth. The error histogram (Figure~\ref{fig:test_error_hist}) is unimodal and centered close to zero, with the majority of errors falling within a narrow range (e.g., $\pm0.015 C_d$). This demonstrates that the model generalizes well to unseen data and does not exhibit significant systemic bias. The MaxAE of 0.045 indicates that even the largest errors are within a reasonable range for early-stage design guidance.

\section{Discussion}\label{sec4}
The results presented in Section~\ref{sec3} demonstrate that our proposed sequential slice-based model achieves high accuracy and efficiency for aerodynamic drag prediction. The $R^2$ value of 0.9528 on the DrivAerNet++ test set signifies that the model captures over 95\% of the variance in drag coefficients, performing comparably to more complex state-of-the-art methods such as TripNet \cite{chen2025tripnet}, while using a simpler architecture and fewer parameters. The low MAE of approximately \(6.046 \times 10^{-3}\) indicates that the average prediction error is very small, making it a reliable tool for shortlisting design candidates.

The effectiveness of the slice-based approach can be attributed to its ability to explicitly model the geometric evolution of the vehicle’s shape along the \textbf{streamwise axis}, a progression that is critical to drag formation. Aerodynamic drag is highly sensitive to how the cross-sectional area and shape change from the front of the vehicle to the rear. By processing an ordered sequence of 2D slices in this direction, the PointNet2D module learns salient features from each local cross-section, and the Bi-LSTM integrates this information to learn the impact of these \textbf{progressive changes}. This sequential analysis is key; slicing in other directions, such as at right angles to the flow, would fail to capture this \textbf{crucial front-to-back narrative}, as each slice would contain parts of the vehicle's front, middle, and rear, thereby losing the causal geometric information beneficial for drag prediction. This limitation is also found in methods that treat the point cloud as an unordered set.

A key advantage of our model is its computational efficiency. With an inference time of $\sim$0.025 seconds per sample on a consumer-grade GPU, it is substantially faster than traditional CFD and many complex deep learning surrogates (e.g., those based on 3D convolutions or full transformers on point clouds). This speed is critical for early-stage design, enabling engineers to rapidly evaluate numerous design variations, perform sensitivity analyses, and conduct shape optimization interactively. Furthermore, the relatively low parameter count contributes to faster training times and reduced risk of overfitting, especially when datasets might be limited for very specific vehicle types not yet covered by large public benchmarks.

Compared to other approaches, our method strikes a balance between performance and interpretability. While global 3D methods like PointNet or DGCNNs can be powerful, their permutation-invariant nature or complex graph structures can make it difficult to understand which parts of the geometry contribute most to the prediction. Our slice-based sequence allows, at least conceptually, for an investigation into how individual slices or segments of slices influence the final $C_d$ prediction through analysis of LSTM activations or attention mechanisms if a transformer were used for sequence modeling. This potential for enhanced interpretability, by linking drag to longitudinal sections, can provide designers with more actionable feedback.

Despite its strengths, the proposed method has limitations.
\begin{enumerate}
    \item \textbf{Inter-slice Information Loss:} While 80 slices provide good resolution, some fine 3D geometric details that do not significantly alter the 2D profile of any single slice but exist between slice planes or are inherently 3D in nature (e.g., complex underbody channels, small winglets with specific orientations not aligned with slices) might not be fully captured. The projection onto the $YZ$ plane also means that 3D curvature within a slice's thickness is lost.
    \item \textbf{Surface-only Scalar Prediction:} The current model predicts only the scalar $C_d$ value and does not provide information about pressure or velocity fields on the vehicle surface or in the surrounding flow. Such field predictions are valuable for detailed aerodynamic analysis and are offered by some more complex surrogate models \cite{chen2025tripnet,kossaifi2025figconv}.
    \item \textbf{Absence of Explicit Physics Priors:} The model is purely data-driven. It does not inherently enforce physical laws like conservation of mass or momentum. This could lead to less robust predictions for out-of-distribution shapes not well-represented in the training data.
    \item \textbf{Fixed Number of Slices:} The choice of $S=80$ slices was based on empirical observation. An adaptive slicing strategy, allocating more slices to regions with higher geometric variation, could improve performance or efficiency but would also introduce additional complexity.

\end{enumerate}

The broader impact of such a fast and accurate surrogate model is the potential to democratize aerodynamic analysis in the early stages of automotive design. It allows for more extensive design space exploration, leading to potentially more aerodynamically efficient vehicles developed in shorter timeframes and at lower costs.

\section{Conclusion}\label{sec5}
This paper introduced a lightweight and efficient neural network architecture for predicting automotive aerodynamic drag coefficients ($C_d$) from 3D vehicle point clouds. Our novel approach transforms the 3D geometry into an ordered sequence of 2D cross-sectional slices along the streamwise axis. These slices are individually encoded using a PointNet2D module, and their sequential geometric evolution is captured by a bi-directional LSTM, with a final MLP regressing the $C_d$.

Trained and evaluated on the large-scale DrivAerNet++ dataset, our model achieved a coefficient of determination ($R^2$) of 0.9528 and a mean absolute error (MAE) of $6.046 \times 10^{-3}$. These results are competitive with more complex state-of-the-art surrogate models, demonstrating the efficacy of the slice-based sequential representation. A key advantage of our method is its computational efficiency, with an inference time of approximately 0.025 seconds per vehicle on a consumer-grade hardware and a modest parameter count of 2.79 million. This enables rapid aerodynamic feedback, facilitating extensive design iteration and optimization in the early conceptual phases of vehicle development. The inherent structure of the model, focusing on longitudinal geometric progression, also offers potential for enhanced interpretability compared to global "black-box" 3D models.

Future work will focus on several avenues. Enhancing slice encoding with more sophisticated 2D shape descriptors or exploring advanced sequence models like transformers could further improve accuracy. Extending the model to predict surface pressure distributions or even simplified flow fields would provide richer aerodynamic insights. Investigating adaptive slicing techniques and incorporating physics-informed neural network principles could enhance robustness and accuracy, particularly for out-of-distribution geometries. Ultimately, integrating such fast and interpretable surrogate models into interactive CAD tools holds the promise of significantly accelerating and improving aerodynamic design in the automotive industry.

\begin{credits}
\subsubsection{\discintname}
The authors have no competing interests to declare that are relevant to the content of this article.
\end{credits}
\bibliographystyle{splncs04}
\bibliography{refs}

\end{document}